\documentclass[letterpaper, 10 pt, conference]{ieeeconf} %

\IEEEoverridecommandlockouts   
\usepackage{amsmath,amsfonts}
\usepackage{algorithmic}
\usepackage{array}
\usepackage{textcomp}
\usepackage{stfloats}
\usepackage{url}
\usepackage{verbatim}
\usepackage{graphicx}  
\usepackage{multirow}  

\usepackage{balance}                                 

\usepackage{amsmath}
\usepackage{algorithmic}
\usepackage{array}
\usepackage{url}

\usepackage{color, array}
\definecolor{MyColor}{gray}{0.75}
\usepackage{colortbl}

\usepackage{graphics} 
\usepackage{epsfig} 
\usepackage{xcolor}

\usepackage{amssymb}
\usepackage{adjustbox}
\usepackage{booktabs} 
\usepackage[ruled,vlined]{algorithm2e}
\usepackage{hyperref}
\usepackage{graphicx}
\usepackage{caption}
\usepackage{amsmath,amssymb, fancyhdr, color, comment, graphicx, environ}
\usepackage{xcolor}
\usepackage{mdframed}
\usepackage{indentfirst}
\usepackage{subcaption}
\usepackage{wrapfig}
\usepackage{ltablex}
\usepackage{siunitx}
\usepackage{caption}
\newcolumntype{L}{>{\raggedright\arraybackslash}X}

\newcommand{\CP}[1]{\ignorespaces}
\newcommand{\ie}{\textit{i.e.}}
\newcommand{\eg}{\textit{e.g.}}

\def  \figsize {0.158} 
\def  \leftvertical {0.8} 
\def  \bottom {1.5} 
\def  \top {2.5} 
\def \textsize {\scriptsize}

\colorlet{first}{gray!30}
\colorlet{second}{gray!15}
\colorlet{third}{gray!7}

\colorlet{table}{gray!20}

\begin{document}


\title{\LARGE \bf SPVSoAP3D: A Second-order Average Pooling Approach to enhance 3D Place Recognition in Horticultural Environments}

\author{T. Barros$^{1}$, C. Premebida$^{1}$, S. Aravecchia$^{2}$, C. Pradalier$^{2}$, U.J. Nunes$^{1}$%
\thanks{$^{1}$T.Barros, C.Premebida and U.J.Nunes are with the University of Coimbra, Institute of Systems and Robotics, Department of Electrical and Computer Engineering, Portugal.
E-mails:{\tt\small\{tiagobarros,~cpremebida,~urbano\}@isr.uc.pt}}%
\thanks{
$^{2}$ IRL2958 GeorgiaTech-CNRS, Metz 57070, France.
}}

\twocolumn[{%
\renewcommand\twocolumn[1][]{#1}%
\maketitle
\begin{center}
    \centering
    \captionsetup{type=figure}
    \includegraphics[width=\textwidth]{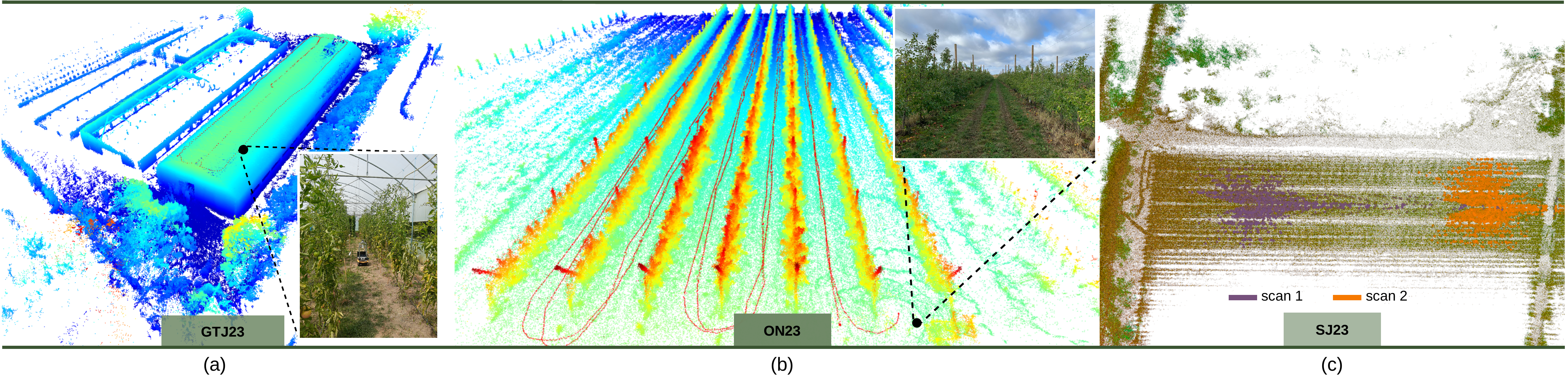}
    
    \captionof{figure}{3D maps of the two new sequences from horticultural environments proposed in this work are as follows: (a) GTJ23, recorded in a greenhouse tomato production facility in Coimbra, Portugal, and (b) ON23, recorded in an orchard in Metz, France. Additionally, (c) illustrates a 3D map of an HORTO-3DLM sequence (SJ23) with two scans, showcasing the permeable nature of these environments to LiDAR scans.} %
    \label{fig:front}
\end{center}%
}]

\renewcommand{\thefootnote}{\fnsymbol{footnote}}
\footnotetext[0]{\vspace{0cm}$^1$ T. Barros, C. Premebida and U.J.Nunes are with the University of Coimbra, Institute of Systems and Robotics, Department of Electrical and Computer Engineering, Portugal.
E-mails: {\tt\footnotesize \emph\{tiagobarros,~cpremebida,~urbano\}@isr.uc.pt}}
\footnotetext[0]{\vspace{0cm}$^2$ S. Aravecchia and C. Pradalier are with the IRL2958 GeorgiaTech-CNRS, Metz 57070, France. E-mail: {\tt\footnotesize \emph \{stephanie.aravecchia, cedric.pradalier\}@georgiatech-metz.fr}} %

%

\begin{abstract}


3D LiDAR-based place recognition has been extensively researched in urban environments, yet it remains underexplored in agricultural settings. Unlike urban contexts, horticultural environments, characterized by their permeability to laser beams, result in sparse and overlapping LiDAR scans with suboptimal geometries. This phenomenon leads to intra- and inter-row descriptor ambiguity.
In this work, we address this challenge by introducing SPVSoAP3D, a novel modeling approach that combines a voxel-based feature extraction network with an aggregation technique based on a second-order average pooling operator, complemented by a descriptor enhancement stage. Furthermore, we augment the existing HORTO-3DLM dataset by introducing two new sequences derived from horticultural environments. We evaluate the performance of SPVSoAP3D against state-of-the-art (SOTA) models, including OverlapTransformer, PointNetVLAD, and LOGG3D-Net, utilizing a cross-validation protocol on both the newly introduced sequences and the existing HORTO-3DLM dataset.
The findings indicate that the average operator is more suitable for horticultural environments compared to the max operator and other first-order pooling techniques. Additionally, the results highlight the improvements brought by the descriptor enhancement stage. The code is publicly available at \hyperlink{https://github.com/Cybonic/SPVSoAP3D.git}{https://github.com/Cybonic/SPVSoAP3D.git}
\end{abstract}

\section{INTRODUCTION}

Autonomous robots are pivotal in enhancing productivity within the agricultural sector \cite{sparrow2021robots, rivera2023lidar}. To operate reliably, these robots require precise and accurate localization systems. In GNSS-denied environments, perception-based methodologies such as Simultaneous Localization and Mapping (SLAM) and place recognition offer viable solutions. Among these, 3D LiDAR-based place recognition has gained prominence both as a component of SLAM and as a standalone method, providing an efficient approach for global localization and loop detection \cite{7747236}.

3D LiDAR-based place recognition has primarily been advanced by the autonomous driving community, which has introduced novel methodologies specifically designed for urban-like environments \cite{barros2021place}. Recent efforts have focused on adapting these methods to natural and agricultural environments \cite{guo2019local, cheng2023treescope}, which pose fundamentally different challenges. Unlike urban settings that return scans with well-defined geometries (e.g., corners, planes, lines), agricultural, and particularly horticultural environments, produce LiDAR scans with poor geometries. The permeable nature of these environments to laser beams allows the beams to traverse various row layers, resulting in scans that cover multiple rows (illustrated in Fig \ref{fig:front}.c). Consequently, this generates sparse scans with poor geometries and highly overlapped scans from neighboring rows, leading to intra- and inter-row descriptor ambiguity.

This work addresses the challenge of 3D LiDAR place recognition, specifically focusing on the inter-row descriptor ambiguity within horticultural environments, by making two key contributions. Firstly, we introduce two new curated sequences comprising 3D LiDAR and GNSS/localization data, thereby enhancing the existing HORTO-3DLM dataset\footnote{https://github.com/Cybonic/HORTO-3DLM.git}. These sequences add diversity and representativeness to the dataset; one sequence was recorded in an apple orchard in Metz, France, using a 16-beam 3D LiDAR mounted on a mobile platform, and the other in a greenhouse in Coimbra, Portugal, employing a 64-beam 3D LiDAR on a different mobile platform. Figure \ref{fig:front} illustrates 3D maps of each sequence.

\begin{figure*}[t]
    \centering
    \includegraphics[width=\textwidth, trim={0cm 0cm 0cm 0cm},clip]{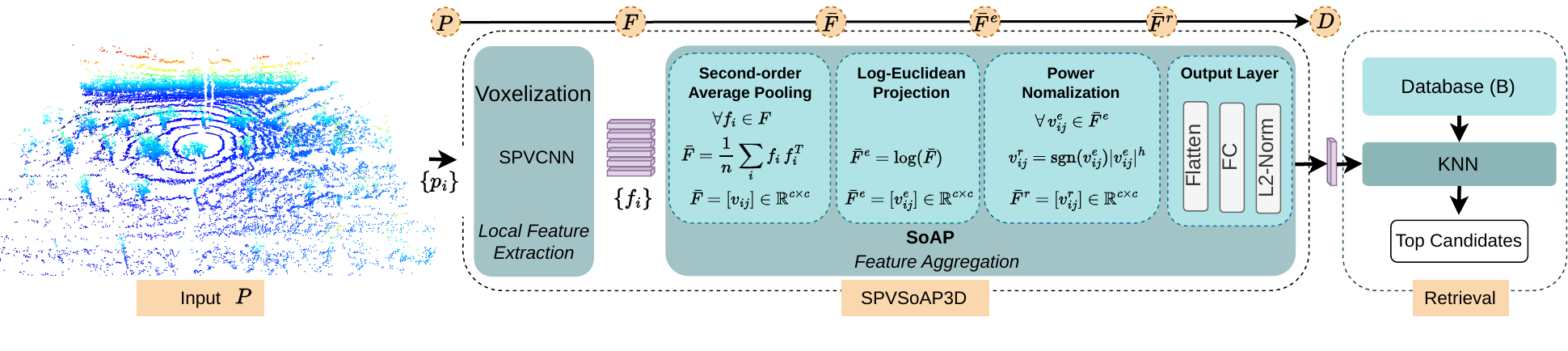}
    \caption{SPVSoAP3D in a retrieval-based framework.  SPVSoAP3D has 5 main stages. In \textbf{stage 1}, the input 3D LiDAR scan $P$ is voxelized and fed to the backbone SPVCNN, which returns local features $F$. In  \textbf{stage 2}, these local features are aggregated using second-order average pooling, resulting in $\bar{F}$. In \textbf{stage 3}, the aggregated features $\bar{F}$ are projected to a tangent space using Log-Euclidean projection, yielding $\bar{F}^e$. 
    In \textbf{stage 4}, the $\bar{F}^e$ features are rescaled using power normalization. In \textbf{stage 5}, the rescaled features $\bar{F}^r$ are flattened and fed to a fully connected layer followed by $L2$-norm. Finally, the model outputs a descriptor $D$, which is used to query the database for the top-k most similar descriptors.}
    \label{fig:pipeline}
\end{figure*}

Secondly, we propose a novel 3D LiDAR-based place recognition method named SPVSoAP3D. SPVSoAP3D uses a sparse point-voxel-based backbone to extract local features from 3D LiDAR scans and utilizes a second-order average pooling approach to aggregate these features into a descriptor. Our contribution lies particularly in the adoption of the average operator as an aggregation approach for 3D LiDAR-based frameworks and a descriptor enhancement approach based on Log-Euclidean projection and power normalization. The proposed approach is illustrated in Fig. \ref{fig:pipeline}.

The proposed approach was evaluated using the HORTO-3DLM dataset, along with two additional sequences, benchmarking SPVSoAP3D against state-of-the-art (SOTA) models such as OverlapTransformer \cite{9785497}, PointNetVLAD \cite{angelina2018pointnetvlad}, and LOGG3D-Net \cite{9811753} using a cross-validation protocol, specifically the \textit{leave-one-out} approach. The empirical results show that average pooling outperforms the second-order max pooling approach used in LOGG3D-Net \cite{9811753}, as well as other first-order pooling methods employed in SOTA 3D LiDAR place recognition models. Additionally, the enhancement stage further improves performance.

\section{RELATED WORK}
Over the past few years, the majority of advancements in 3D LiDAR place recognition have been proposed for urban environments \cite{barros2022attdlnet,siva2020voxel}. These advancements have been mainly driven by Deep Neural Networks (DNNs), which have enabled the learning of features directly from point clouds in an end-to-end fashion \cite{barros2021place}. A less studied field is place recognition in agricultural or natural environments, where efforts have recently been directed towards adapting urban-based approaches for these semi-structured environments. Despite these efforts, such endeavors face significant challenges due to the limited availability of datasets, given the substantial amounts of data required to train DL-based approaches.

\subsection{Datasets}

The autonomous driving community has extensively utilized well-established urban environment datasets to advance 3D LiDAR-based place recognition. Datasets such as KITTI \cite{geiger2013vision}, KITTI360 \cite{liao2022kitti}, NCLT \cite{carlevaris2016university}, Oxford \cite{maddern20171}, and MulRan \cite{jeong2019complex} are just a few examples of popular benchmarks used in both SLAM and place recognition tasks. The diversity in these datasets in terms of environment and sensor characteristics (e.g., resolution, range, or precision) make them  well-suited for training and evaluating DL-based place recognition approaches.

In contrast, natural and agricultural environments possess different characteristics compared to on-road urban environments. In recent years, new LiDAR datasets from these environments have emerged to bridge this gap. Datasets such as Wild-places \cite{knights2023wild}, RELLIS-3D \cite{jiang2021rellis}, ORFD \cite{min2022orfd}, TreeScope \cite{cheng2023treescope}, and VineLiDAR \cite{prabhu2024uavs} exemplify these efforts to provide multi-modality real-world sensory data of natural  and agricultural environments for localization, mapping, and particularly for place recognition tasks.

Despite these efforts, there remains a lack of representative data from environments such as horticulture. Therefore, one of the objectives of this work is to provide real-world sensory data from such environments, specifically from two distinct settings: an apple orchard and a tomato plantation within a greenhouse. These two sequences will complement an existing horticulture dataset called HORTO-3DLM, which includes 3D LiDAR sensory data from orchards of apples and strawberries.

\subsection{3D LiDAR-based Place Recognition}

Although hand-crafted descriptors are still used, Deep Learning (DL) techniques have become increasingly popular for place modeling \cite{barros2021place}. Among these techniques, most 3D LiDAR-based place recognition approaches rely on DNNs, typically comprising two main modules: a backbone for extracting local features from input point clouds, and a feature aggregation module for combining these local features into a global descriptor.

In this work, we focus on the feature aggregation module, an area extensively explored in vision perception tasks such as classification \cite{gao2019global,yu2018statistically} and segmentation \cite{carreira2012semantic}, where first-order and second-order statistics are commonly compared. These methods are less studied in 3D LiDAR-based place recognition.  Many works in the field resort to first-order methods such as NetVLAD \cite{angelina2018pointnetvlad,arandjelovic2016netvlad,cattaneo2022lcdnet}, or generalized mean approaches \cite{komorowski2021minkloc3d}, while  higher-order pooling receive less attention. Recent works such as LOGG3D-Net \cite{9811753} and Locus \cite{vidanapathirana2021locus} explore second-order pooling utilizing the max operator to aggregate features.

Our approach builds upon the concept of second-order pooling, proposing the use of the average operator instead of the max operator. We show that this operator serves as a more suitable feature aggregator for second-order features in horticultural environments than the max pooling operator. Our findings align with those in vision-related research \cite{gao2019global,yu2018statistically}, which have shown the superiority of second-order average pooling, for instance, in image segmentation tasks \cite{carreira2012semantic}. Furthermore, inspired by the work in \cite{carreira2012semantic}, we additionally project the average representation into a Log-Euclidean tangent space and apply power normalization. Our results indicate that these additional stages further enhance performance.

\section{PROPOSED APPROACH}
\label{sec:pa}

This section details the proposed SPVSoAP3D model within a retrieval-based place recognition framework as shown in Fig.~\ref{fig:pipeline}. In this work, we frame the place recognition problem as a retrieval tasks, evaluated at the segment level, meaning that the plantations are divided into segments. For instance, each row constitutes a segment, and the extremities of the plantations (perpendicular to the rows) are also considered segments. This segmentation approach enables the evaluation of performance at the segment level.

\subsection{Problem Formulation}
A retrieval-based 3D LiDAR place recognition framework comprises two stages: first, a LiDAR scan  $P \in \mathbb{R}^{n\times 3}$ with $n$ points, which is mapped to a descriptor $D \in \mathbb{R}^{d}$, using a function $\Theta:\mathbb{R}^{n\times 3} \rightarrow \mathbb{R}^{d}$; second, the descriptor $D$ queries a database for the top-k candidates based on a similarity metric. The retrieved candidates represent potential revisited places. The modeling capacity of $\Theta$ directly impacts  the retrieval performance.

This work focuses on the mapping function $\Theta$,  which can be decomposed into two functions, $\Theta \equiv  \varphi \circ \phi$, where $\phi:\mathbb{R}^{n\times 3} \rightarrow  \mathbb{R}^{n \times c}$ maps an input scan $P$ (with 3 dimensions) to a hyper-dimensional feature vector $F = \{f_i\} \in \mathbb{R}^{n \times c}$ with $c$ dimensions; and $\varphi: \mathbb{R}^{n\times c} \rightarrow \mathbb{R}^{d}$ aggregates the features $F$ into a global descriptor $D$ with $d$ dimensions.

\subsection{SPVSoAP3D}
The proposed place modeling approach, SPVSoAP3D, inspired by LOGG3D-Net \cite{9811753}, can be summarized in five main stages: (1) local features extraction; (2) second-order average pooling; (3) Log-Euclidean tangent space projection; (4) power normalization; and (5) linear projection using a fully connected layer (FC) and descriptor normalization. The model is trained using a contrastive learning approach employing a triplet loss for parameter optimization.   

\subsubsection{Local Feature Extraction}
Local features are extracted with the Sparse Point-Voxel (\textbf{SPV}) CNN \cite{tang2020searching}, which serves as a backbone. Initially, the input scan  $P = \{(x,y,z)_i\} \in \mathbb{R}^{n\times 3}$ is converted to a voxel representation $V  \in \mathbb{R}^{n'\times 3}$ with $n'$ occupied voxels where $n'<n$. This voxel representation is then processed by the SPVCNN backbone to extract the local features $F = \{f_i\} \in \mathbb{R}^{n'\times c}$, consisting of $n'$ features, and each with $c$ dimensions. Details on the voxelization process and on SPVCNN can be found in \cite{tang2020searching}.

\subsubsection{Second-order Average Pooling}
Local features are aggregated by averaging over second-order interactions (\eg, outer products) \cite{carreira2012semantic}.  Thus, given a set of local features $F$, the second-order average pooling operation is defined as follows:
\begin{equation}
    \bar{F} = \frac{1}{n'} \sum_{f_i \in F} f_i \, f_i^T,
\end{equation}

\noindent where $\bar{F}\in \mathbb{R}^{c\times c}$ is the aggregated average representation.

\subsubsection{Log-Euclidean Projection}
In this work, the descriptors (output of the model) are compared using the $L_2$-norm, which is an Euclidean metric.  However,  $\bar{F}$ leads to a symmetric positive definite (SPD) matrix, forming a Riemannian manifold \ie, a non-Euclidean space \cite{carreira2012semantic,arsigny2007geometric,bahtia2007positive}. The geometric relationships in the manifold can be preserved by mapping the SPDs to a Euclidean tangent space. This mapping can be achieved using a principal matrix logarithm
operation \cite{carreira2012semantic} as follows,
\begin{equation}
     \bar{F}^{e} = \log(\bar{F})  \in \mathbb{R}^{c\times c}.
\end{equation}
The  matrix logarithm $\log(\cdot)$ is implemented as described in \cite{huang2017riemannian,gao2020learning}, it is obtained by computing the logarithm of the eigenvalues \textit{s.t.} $\log(\bar{F})= U\,\log(\Sigma)\, V^T$ with $\Sigma$ being the diagonal matrix containing the eigenvalues of $\bar{F}$. The eigenvalues are computed using the SVD decomposition.

\subsubsection{Power Normalization}
The descriptiveness of the features can be improved reducing the sparsity\,\cite{carreira2012semantic}. One approach to reducing sparsity is through power normalization, which rescales feature values by amplifying smaller values, while saturating larger ones \cite{perronnin2010improving}. Given $\bar{F}^{e} = [ v_{ij}] \in \mathbb{R}^{c\times c}$, the rescaling is computed using element-wise power normalization as follows:
\begin{equation}
    v^r_{ij} = \text{sgn}(v_{ij})|v_{ij}|^h, \ \forall\, v_{ij} \in \bar{F}^{e}  
\end{equation}
\noindent where $v^r_{ij}$ is the rescaled feature element. The rescaled representation is $\bar{F}^r = [v^r_{ij}] \in \mathbb{R}^{c\times c}$, and  $h \in [0,1]$ is the power parameter.
In contrast to previous works such as \cite{carreira2012semantic,perronnin2010improving}, where $h$ is fixed, here $h$ is a trainable parameter. 

\subsubsection{Output Layer}
Before obtaining the descriptor $D \in \mathbb{R}^d$ with $d$ dimensions, $\Bar{F}^r \in \mathbb{R}^{c\times c}$ is flattened and fed to a fully-connected layer $FC: \mathbb{R}^{c^2} \rightarrow  \mathbb{R}^d$, and normalized using the $L_2$-norm.

\subsection{Network Training} \label{sec:nt}

SPVSoAP3D is trained using the LazyTriplet loss as proposed in\,\cite{angelina2018pointnetvlad}, which employs tuples of scans. Given a tuple $\tau_i = (P_a, P_p, \{P_{n_i}\})$, $P_a$ and $P_p$ form an anchor-positive pair, and $\{P_{n_i}\}_{i=1}^m$ is a set of negative scans. 

\begin{figure}[t]
    \centering
    \includegraphics[width=\columnwidth, trim={0cm 0cm 0cm 0cm},clip]{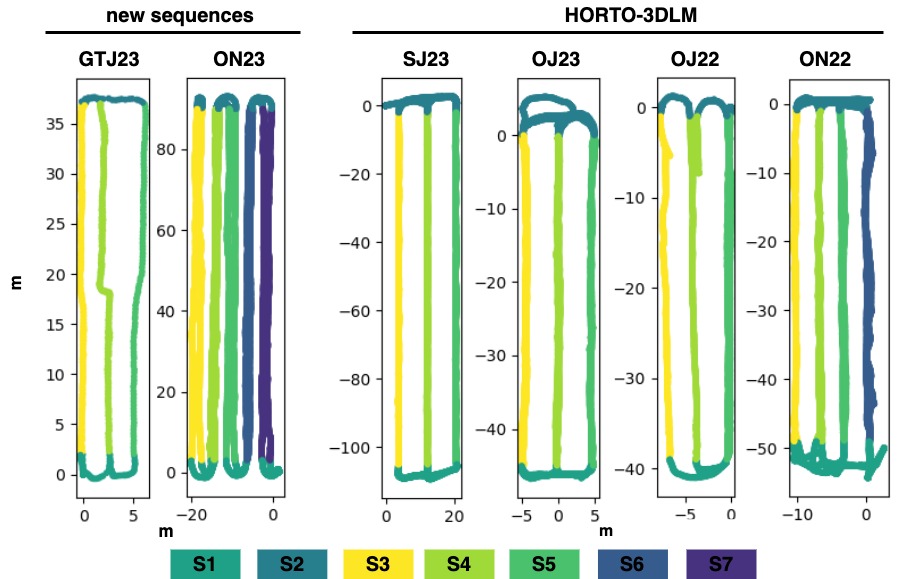}
    \caption{Paths with segments identified by colored S1,...,S7. GTJ3, SJ23, OJ23 and OJ22 have 5 segments, ON22 has 6 segments, while ON23 has 7 segments.}
    \label{fig:rows}
\end{figure}

As in traditional place recognition, the anchor-positive pair must come from the same place. However, in this work, we define "same place" differently by adding a third constraint. Thus, the anchor-positive pair selection is based on the following three constraints (where the third is our additional constraint): (1) both scans must be within a threshold range $r_{th}$ (in meters); (2) both scans must be from different revisits, meaning that the positive scan cannot be the immediate previous scan of the anchor; (3) both scans must be from the same segment (see Fig.~\ref{fig:rows}). All scans that do not satisfy these three constraints are considered negatives.

To compute the loss, $\tau_i = (P_a, P_p, \{P_{n_i}\})$ is mapped to the respective descriptors $\hat{\tau_i} = (D_a, D_p, \{D_{n_i}\})$, which are employed in the loss as follows:
\begin{equation}
    \mathcal{L}_T = \max(d_{AP} - d_{AN} + m, 0),
    \label{eq:lt}
\end{equation}

\noindent where $d_{AP} = \|D_a - D_p\|_2$ is the Euclidean distance between the anchor and the positive, and $d_{AN} = \min\limits_{i} (\|D_a - \{D_{n_i}\}\|_2)$ is the Euclidean distance between the anchor and the hardest negative, \ie, the negative that is closest in the descriptor space. Finally, $m$ designates a margin value.

\begin{figure}[t]
    \centering
    \includegraphics[width=\columnwidth, trim={0cm 0cm 0cm 0cm},clip]{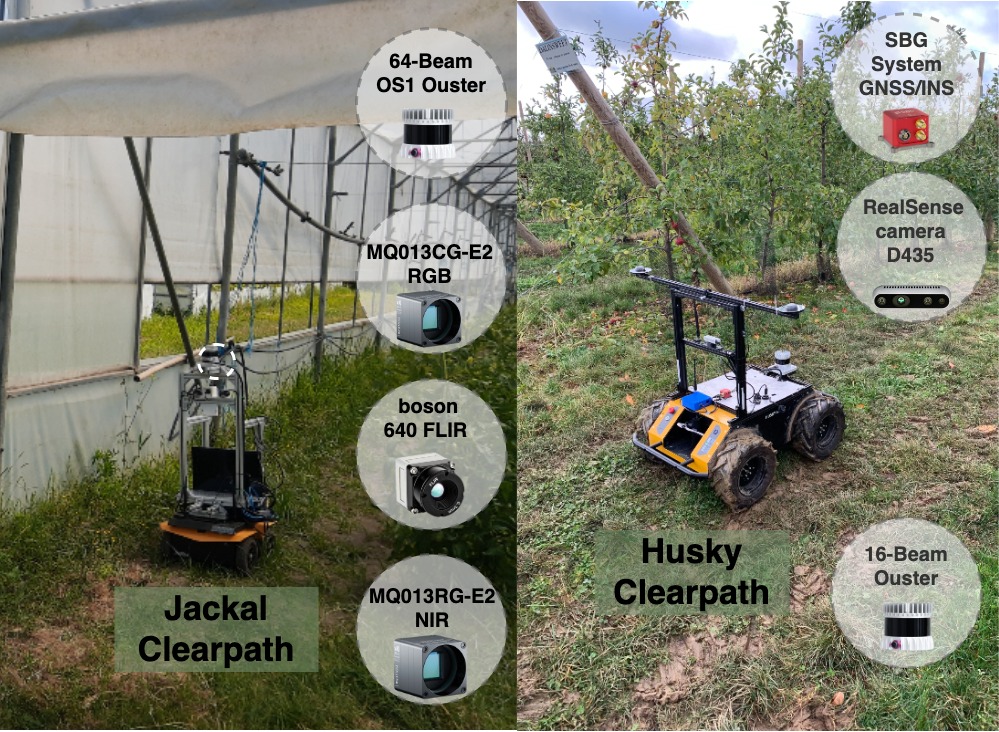}
    \caption{Mobile platforms and sensors used for recording the sequences. The Jackal platform was used to record the GTJ23, while the Husky platform was used to collect data for the ON23 dataset.}
    \label{fig:robots}
\end{figure}

\section{EXPERIMENTAL EVALUATION} \label{sec:experiments}

The proposed approach was evaluated on data from horticultural environments, utilizing two new sequences, as well as the HORTO-3DLM dataset.

\subsection{HORTO-3DLM+}

The HORTO-3DLM dataset comprises four sequences from distinct horticultural environments: three from orchards (two from apples and one from cherries), and another sequence from strawberries within polytunnels with a table-top growing system.  These sequences were recorded in England, UK, using a Clearpath Husky mobile robot equipped with a Velodyne VLP32 3D LiDAR (10Hz) and a ZED-F9P RTK-GPS (5Hz).  The objective of this dataset is to provide real-world sensory data from horticultural environments to support research, particularly in localization, mapping, and place recognition.

In this work, we introduce two additional sequences to enhance the original HORTO-3DLM: one sequence recorded in an apple orchard in Metz, France (referred to as ON23); and another recorded  in a greenhouse tomato production in Coimbra, Portugal (referred to as GTJ23). Each sequence was recorded with a different setup, illustrated in Fig. \ref{fig:robots}. 

Sequence ON23 was captured in November of 2023 with an  16-beam Ouster 3D LiDAR and an SBG GNSS/INS system (without RTK) mounted on a Clearpath Husky mobile platform.  To address the low LiDAR resolution, the original scans were merged to increase point density, resulting in sub-maps with approximately 100k points per sub-map. This operation reduced the original sequence from 25836 scans to 3086 sub-maps in total. Both SLAM and GNSS positioning data are provided, and the synchronization between the sub-maps and the positioning data was achieved by associating the nearest neighbor timestamps in each modality, using the first scan in the sub-maps as the timestamp reference.

On the other hand, sequence GTJ23 was recorded in June of 2023 using a 64-beam Ouster 3D LiDAR mounted on a Clearpath Jackal mobile platform. In this sequence the the ground-truth positions were computed using a SLAM approach due to signal interference caused by the greenhouse structure.

Figure \ref{fig:front} outlines the 3D maps of each sequence and their respective environments, while Fig. \ref{fig:robots} illustrates the two mobile platforms and the sensors used in the recordings. The details of the augmented dataset HORTO-3DLM+, which comprises the original four sequences and the two new sequences, are presented in Table \ref{tab:sequence}, while  Fig. \ref{fig:rows} outlines the paths of all six sequences.

\begin{table}[t]
  \centering
  \caption{HORTO-3DLM+: summary of the sequences used in the experiments. The upper four rows represent Seqs. from the  HORTO-3DLM dataset, while the bottom two rows represent the two new Seqs. The `Seq.' column contains the sequence names. The `M', `Y', and `C' columns refer to the month, year, and country of recording, respectively. The total distance (`Dist') of each sequence is measured in meters, while the scan size 
  is given by the number of points in each scan.}
  {\renewcommand{\arraystretch}{1.3}
    \begin{adjustbox}{max width=\linewidth}
        \begin{tabular}{
        >{\centering\arraybackslash}p{0.06\columnwidth}
        >{\centering\arraybackslash}p{0.05\columnwidth}
        >{\centering\arraybackslash}p{0.04\columnwidth}
        >{\centering}p{0.04\columnwidth}
        >{\centering}p{0.05\columnwidth}
        >{\centering\arraybackslash}p{0.05\columnwidth}
        >{\centering\arraybackslash}p{0.04\columnwidth}
        >{\centering\arraybackslash}p{0.08\columnwidth}
        >{\raggedright\arraybackslash}p{0.2\columnwidth}
        }
        \noalign{\hrule height 1pt}\hline	
        Seq. & M & Y &  C & Nº Scans & Nº Rows & Dist. [m] & Scan Size & Plantation Type\\
        \midrule
		\midrule
        ON22 & Nov. & 2022 &  UK & 7974  & 4  & 514 &  48k & Apple (open)\\
        OJ22 & July & 2022 &  UK & 4361  & 3  & 206  & 50k & Apple (open)\\
        OJ23 & June & 2023 &  UK & 7229  & 3  & 459  & 46k & Cherry (open)\\ 
        SJ23 & June & 2023 &  UK & 6389  & 3  & 742 &  48k &Strawberry (polytunnels) \\  \rowcolor{gray!20} \hline 
        ON23 & Nov. & 2023 &  FR & 3086  & 5  & 966& 105k &Apple (open) \\   \rowcolor{gray!20}
        GTJ23 & June & 2023 & PT & 661  & 3  & 202 & 60k &  Tomato (greenhouse)\\ 
       \noalign{\hrule height 1pt}\hline	
        \end{tabular}%
    \end{adjustbox}
    }
  \label{tab:sequence}%
\end{table}%

\begin{table}[t]
  \centering
  \caption{Dataset split used for the evaluation. In leave-one-out cross-validation, a model that is evaluated on a sequence, is trained on the remaining sequences. For instance, to assess a model's performance on sequence OJ22, the same model is trained on OJ23, ON22, SJ23, ON23, and GTJ23.}
        {\renewcommand{\arraystretch}{1}
        \begin{tabular}{>{\centering\arraybackslash}p{0.1\columnwidth}
                        >{\centering\arraybackslash}p{0.25\columnwidth}
                        >{\centering\arraybackslash}p{0.2\columnwidth}
                        >{\centering\arraybackslash}p{0.25\columnwidth}}
        \noalign{\hrule height 1pt}\hline
        \addlinespace[0.1cm]
        & & \multicolumn{2}{c}{Cross-validation}\\\cline{3-4}
        \addlinespace[0.5em]
        \addlinespace[0.1em] \hline
        \addlinespace[0.1cm]
        Sequence  & Training anchors in the Seq. & Test anchors in the Seq.& Remaining training anchors combined \\
        \midrule \midrule
        OJ22  & 100 &  1797 & 2302 \\ 
        OJ23  & 512  & 4801 & 1890 \\
        ON22  & 495  & 7367 & 1907 \\
        SJ23  & 598  & 3509 & 1804 \\  \rowcolor{gray!20} \hline
        ON23  & 580  & 1657 & 1822 \\  \rowcolor{gray!20}
        GTJ23  & 117  & 295 & 2285 \\ 
        \noalign{\hrule height 1pt}\hline	 
    \end{tabular}%
    }
    \label{tab:experiments}%
\end{table}%

\subsection{Dataset and Evaluation Protocol} \label{sec:dep}

In this work, we divide the path into various segments (S1, S2, ..., S7), as illustrated in Fig. \ref{fig:rows}. In addition to the rows, each corresponding to a segment, we also consider the extremities (i.e., S1 and S7) as separate segments. This segmentation allows for the imposition of the third constraint defined in Section \ref{sec:nt}. 

The training set is generated based on the three constraints defined in Section \ref{sec:nt},  using a search radius of $2\,m$ (\ie, $r_{th}=2\,m$) for the anchor-positive pairs. Among all anchor-positive pairs, only the anchors that are at least $0.5$\,m apart are selected to reduce repetitive scans. For each anchor-positive pair, a set of negative scans is randomly selected from the pool of scans that meet two criteria: (1) they originate from different segments, and (2) they are located outside a $10$-meter radius relative to the anchor.

For the evaluation, we employ a cross-validation strategy, specifically the \textit{leave-one-out} protocol, and report the results from two distinct experiments. In the first experiment, we evaluate the model's performance using a fixed radius $r_{th} = 10\,m$ and report the recall@$k$ with $k \in [1, 1\%]$. In the second experiment, we analyze the performance along the segments, \ie, $r_{th} \in [1,...,l]$, where $l$ represents the segment length, and report recall@$k$ with $k \in [1, 10]$.  Table \ref{tab:experiments} presents the training and evaluation data generated with the aforementioned conditions. 

\subsection{Implementation and Training Details}

All models are trained and evaluated under identical conditions. Specifically, all scans are downsampled to 10000 points without cropping.  For the proposed model, the downsampled scans are voxelized with a grid size of $0.1m$. The backbone is configured to return features with 16 dimensions (i.e., $c=16$), and $h$ is initialized at 0.75, while the descriptor size is set to 256 dimensions (i.e., $d=256$).

Regarding training, all models are trained for 50 epochs, using training tuples consisting of an anchor, the closest positive, and 20 negatives. The margin of the loss is set to 0.5 ($m=0.5$), and model parameters are optimized using the AdamW optimizer with a learning rate of 0.0001 and a weight decay of 0.0005. Additionally, all state-of-the-art models are implemented based on the original code.

The models are executed on a machine equipped with an AMD Ryzen 9 5900X 12-Core CPU, 64 GB RAM, and a NVIDIA GeForce RTX 3090 GPU. The code is implemented in Python 3.9, utilizing PyTorch with CUDA 11.7.

\begin{figure*}[h]
 \centering
    \begin{subfigure}[b]{0.02\textwidth}
        \rotatebox{90}{\textsize Recall@1}
        \vspace{30pt}
    \end{subfigure}%
  \begin{subfigure}[b]{\figsize\textwidth}
     \textsize ON23
    \centering
    \includegraphics[width=\textwidth, trim={\leftvertical cm \bottom cm 1.cm \top cm},clip]{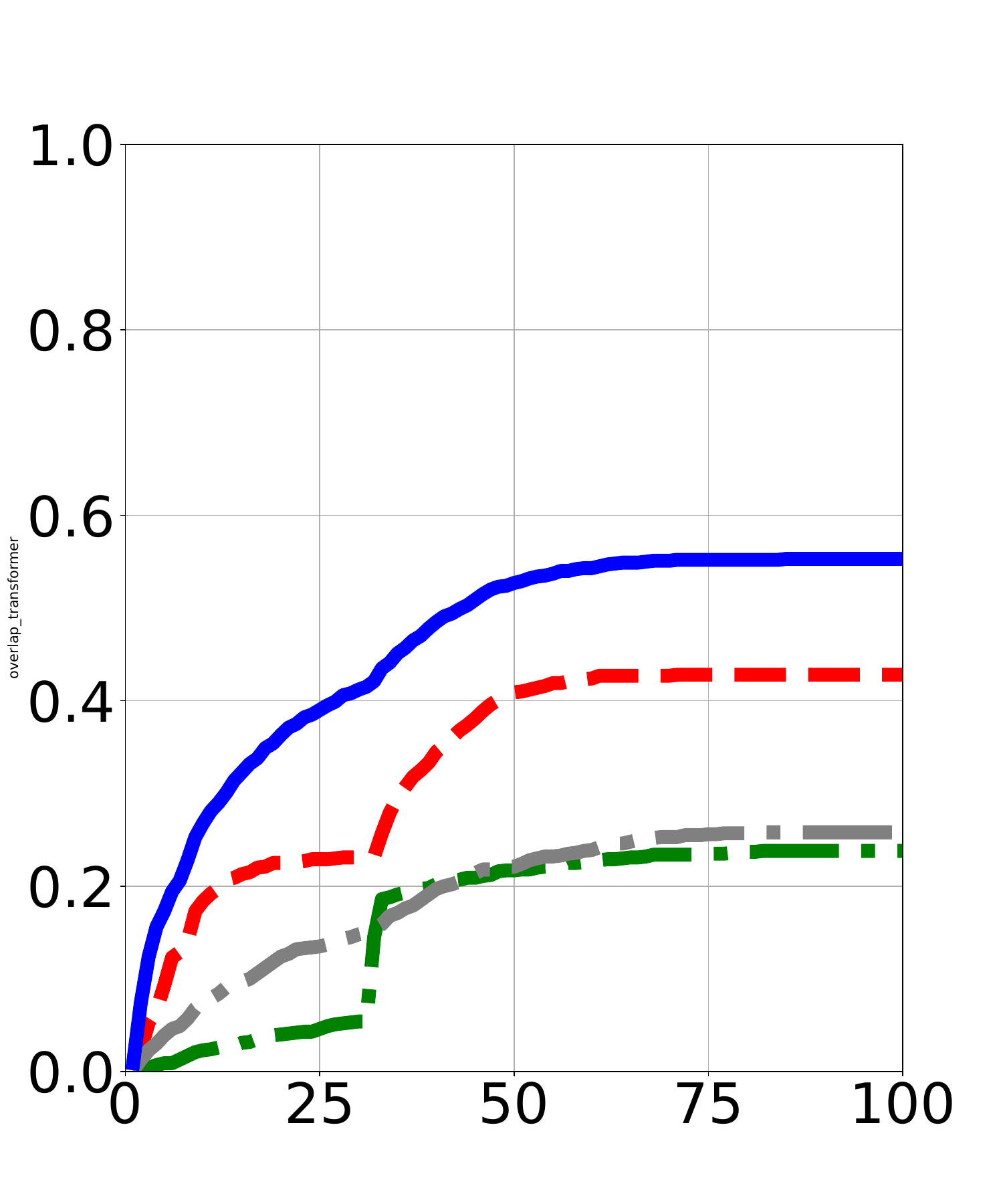}
  \end{subfigure}
  \hfill
  \begin{subfigure}[b]{\figsize\textwidth}
    \textsize GTJ23
    \centering
    \includegraphics[width=\textwidth, trim={\leftvertical cm \bottom cm 1cm \top cm},clip]{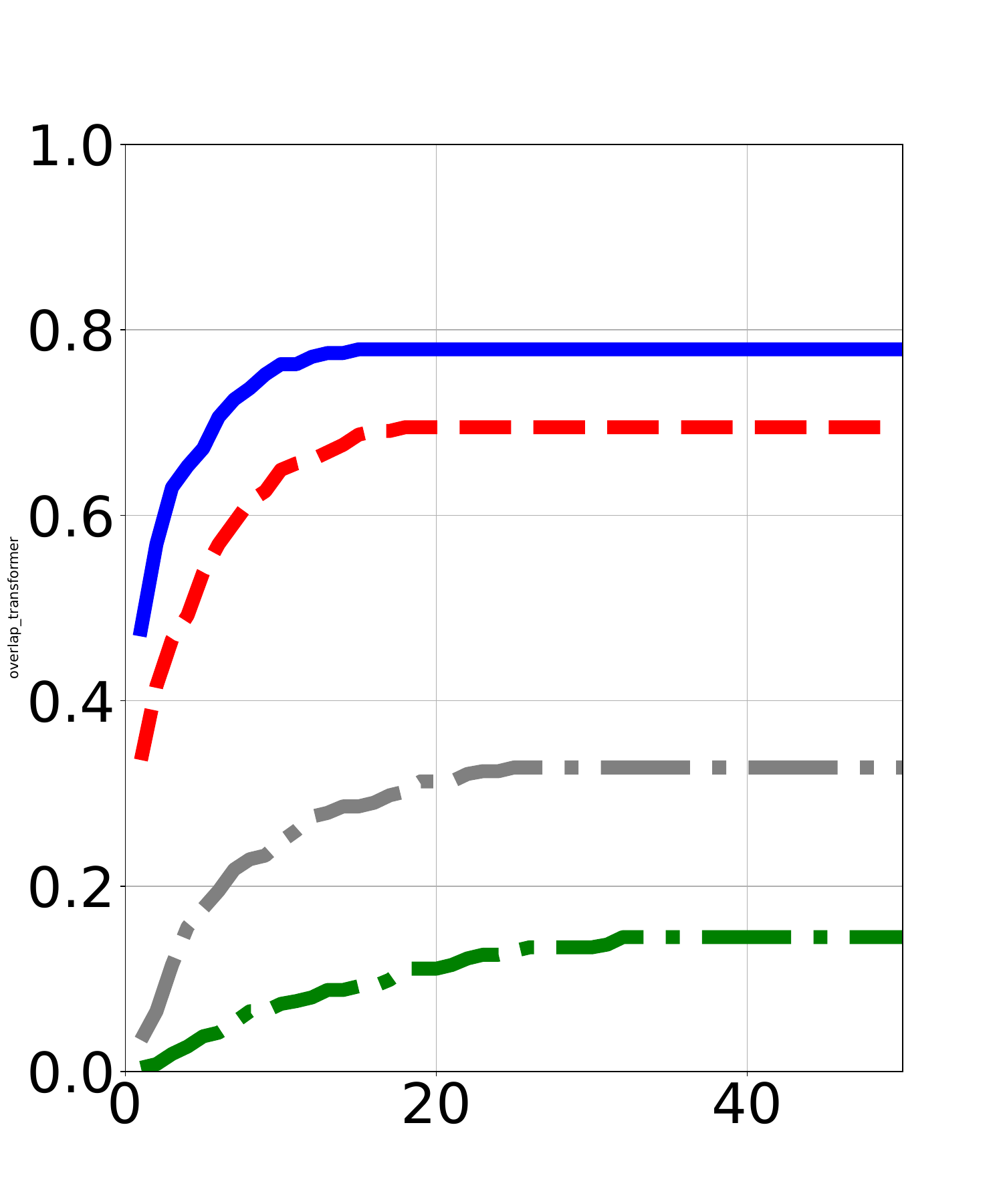}
  \end{subfigure}
  \hfill
  \begin{subfigure}[b]{\figsize\textwidth}
   \textsize ON22
    \centering
    \includegraphics[width=\textwidth, trim={\leftvertical cm \bottom cm 1cm \top cm},clip]{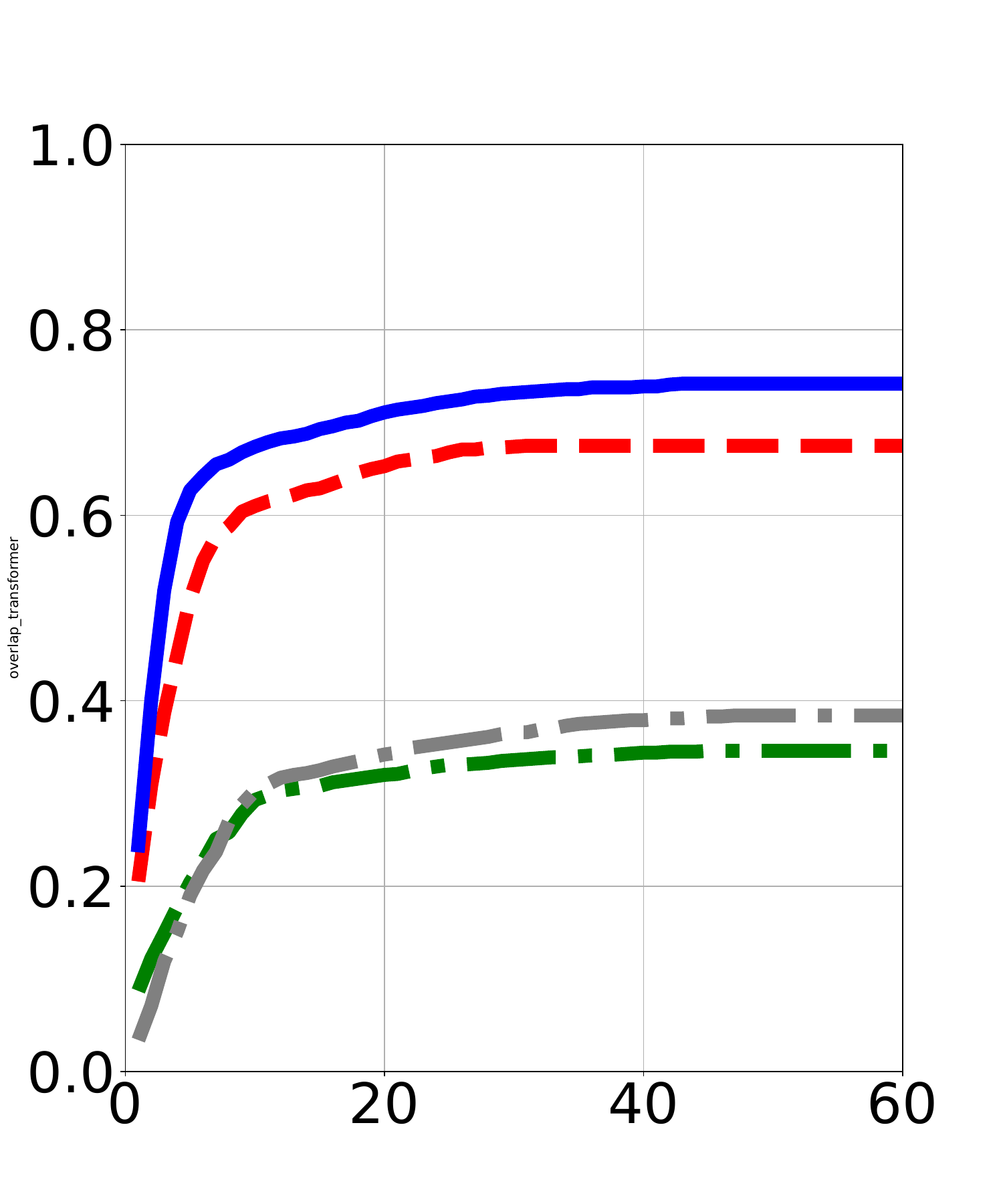}
  \end{subfigure}
  \hfill
  \begin{subfigure}[b]{\figsize\textwidth}
   \textsize  OJ23
    \centering
    \includegraphics[width=\textwidth, trim={\leftvertical cm \bottom cm 1cm \top cm},clip]{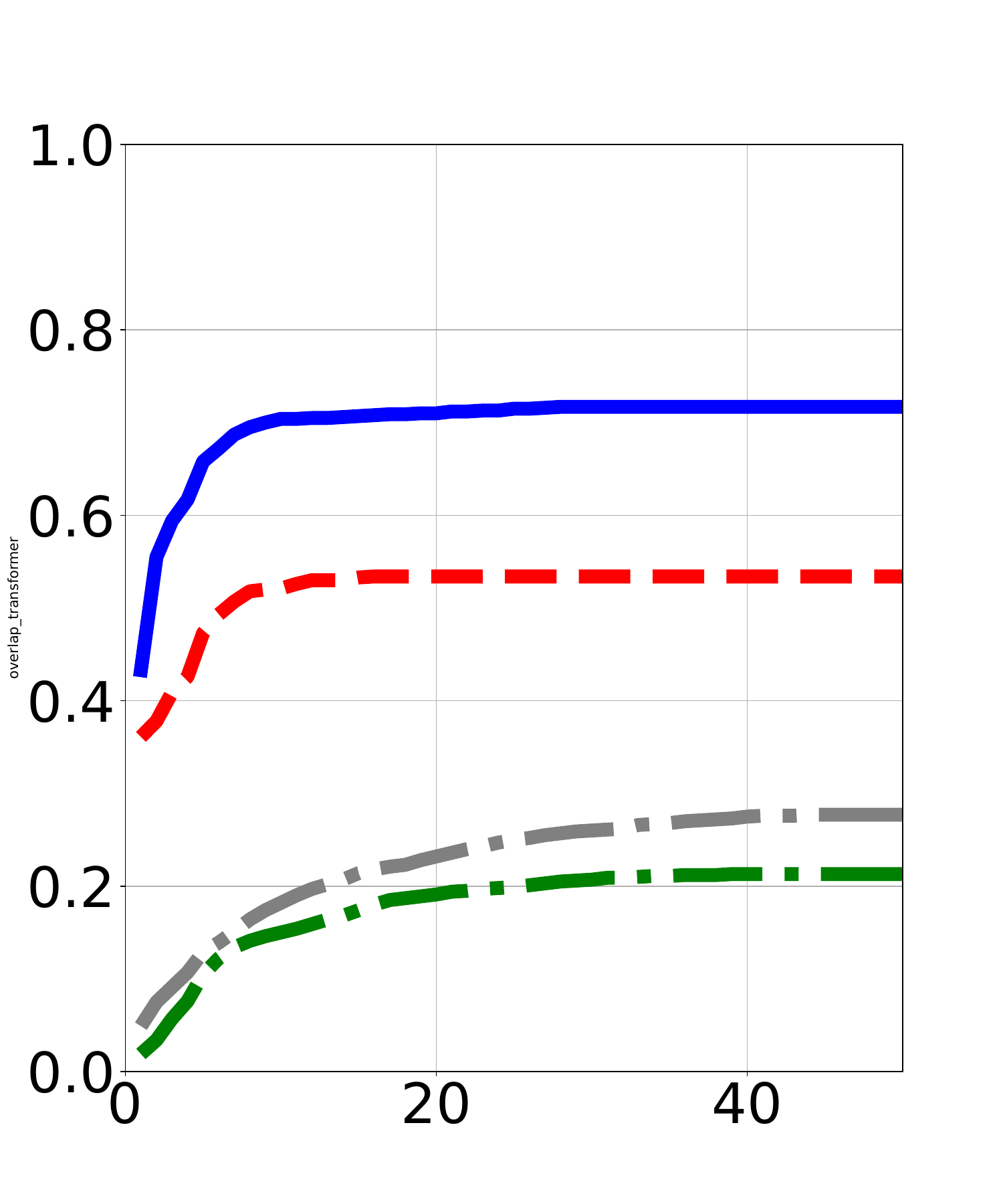}
  \end{subfigure}
  \begin{subfigure}[b]{\figsize \textwidth}
  \textsize  OJ22
    \centering
    \includegraphics[width=\textwidth, trim={\leftvertical cm \bottom cm 1cm \top cm},clip]{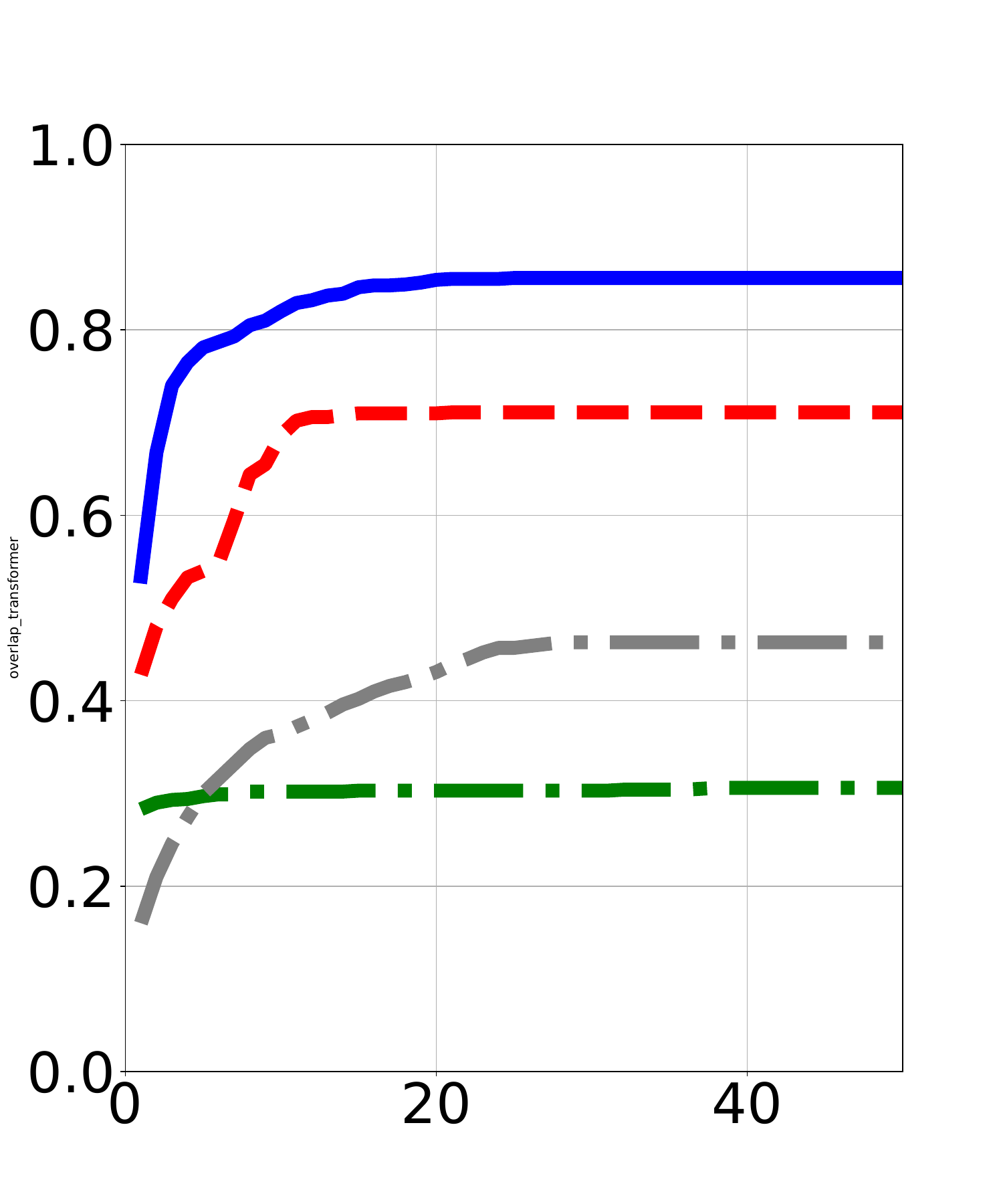}
  \end{subfigure}
  \hfill
  \begin{subfigure}[b]{\figsize\textwidth}
 \textsize SJ23
    \centering
    \includegraphics[width=\textwidth, trim={\leftvertical cm \bottom cm 1cm \top cm},clip]{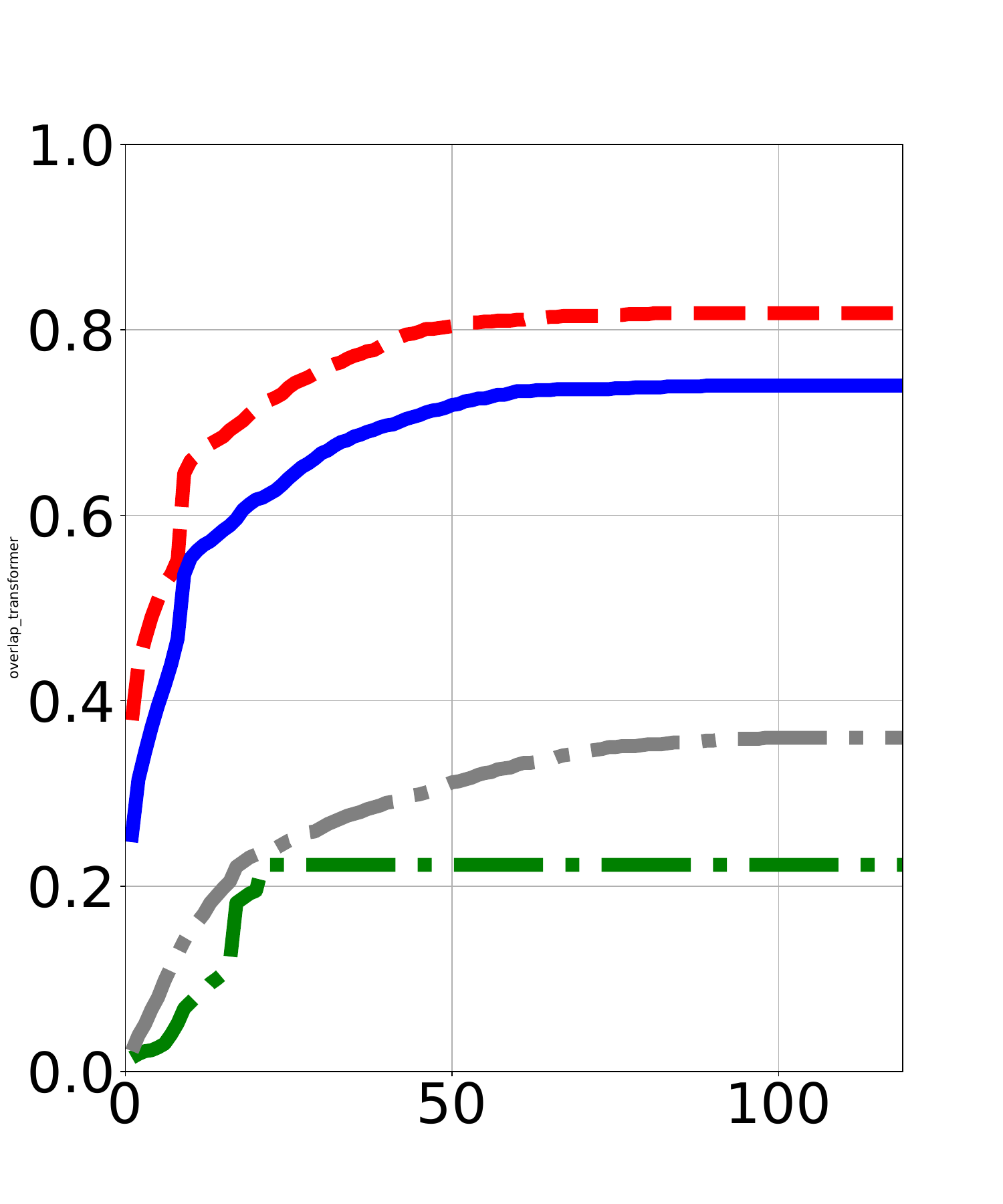}
  \end{subfigure}


  \begin{subfigure}[b]{0.02\textwidth}
        \rotatebox{90}{\textsize  Recall@10}
        \vspace{30pt}
    \end{subfigure}%
  \begin{subfigure}[b]{\figsize\textwidth}
    \centering
    \includegraphics[width=\textwidth, trim={\leftvertical cm \bottom cm 1cm \top cm},clip]{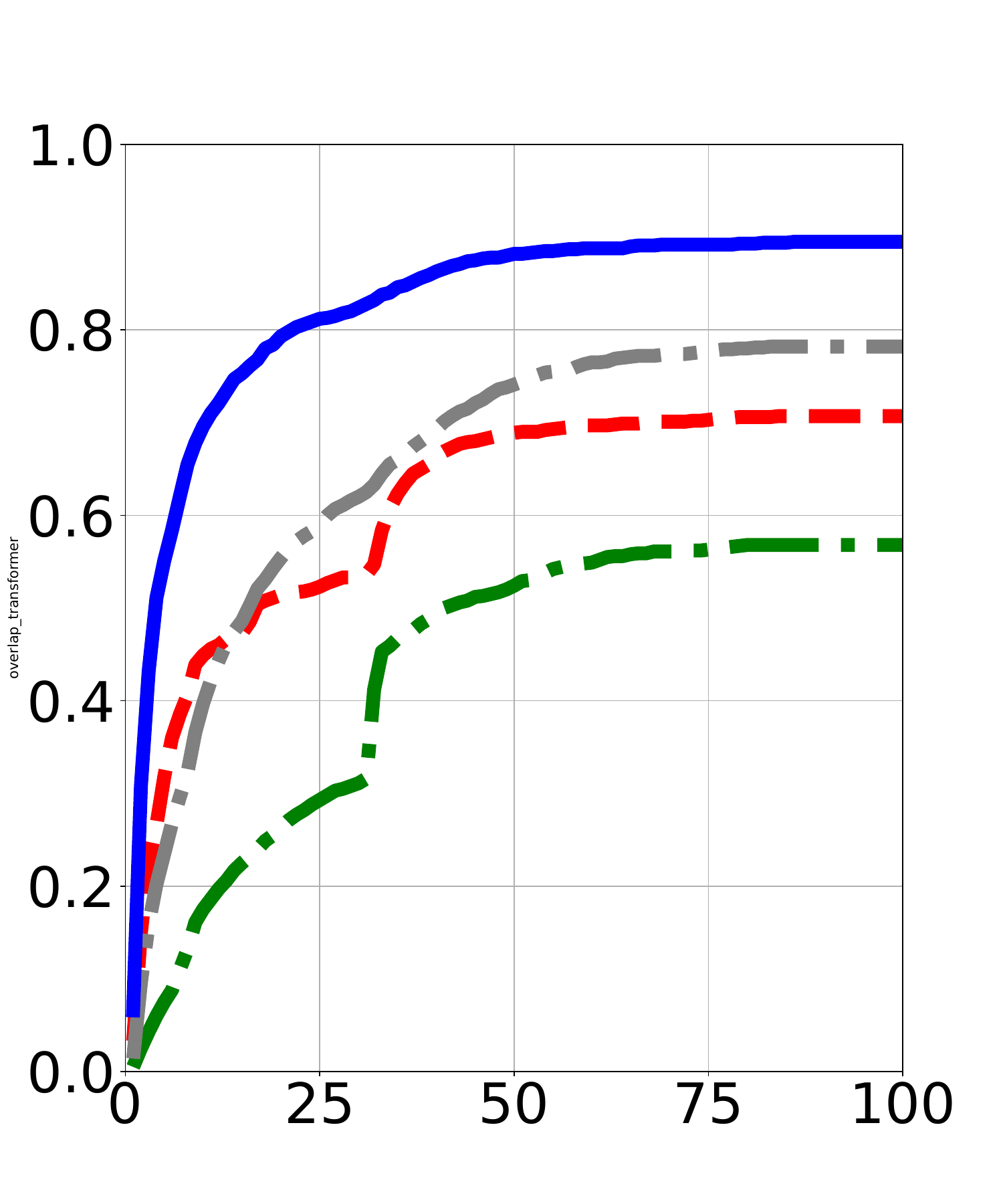}
  \end{subfigure}
  \hfill
  \begin{subfigure}[b]{\figsize\textwidth}
    \centering
    \includegraphics[width=\textwidth, trim={\leftvertical cm \bottom cm 1cm \top cm},clip]{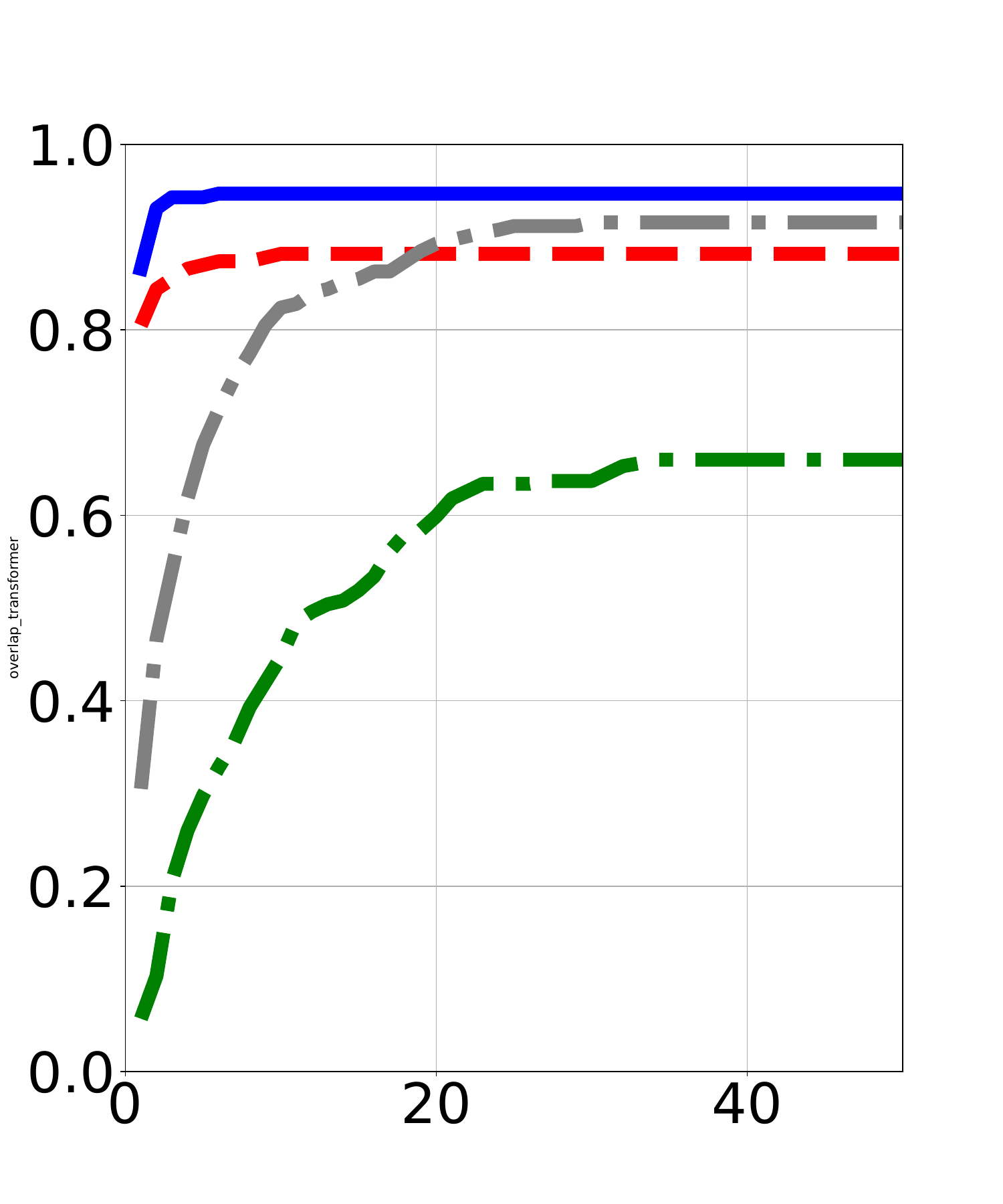}
  \end{subfigure}
  \hfill
  \begin{subfigure}[b]{\figsize\textwidth}
    \centering
    \includegraphics[width=\textwidth, trim={\leftvertical cm \bottom cm 1cm \top cm},clip]{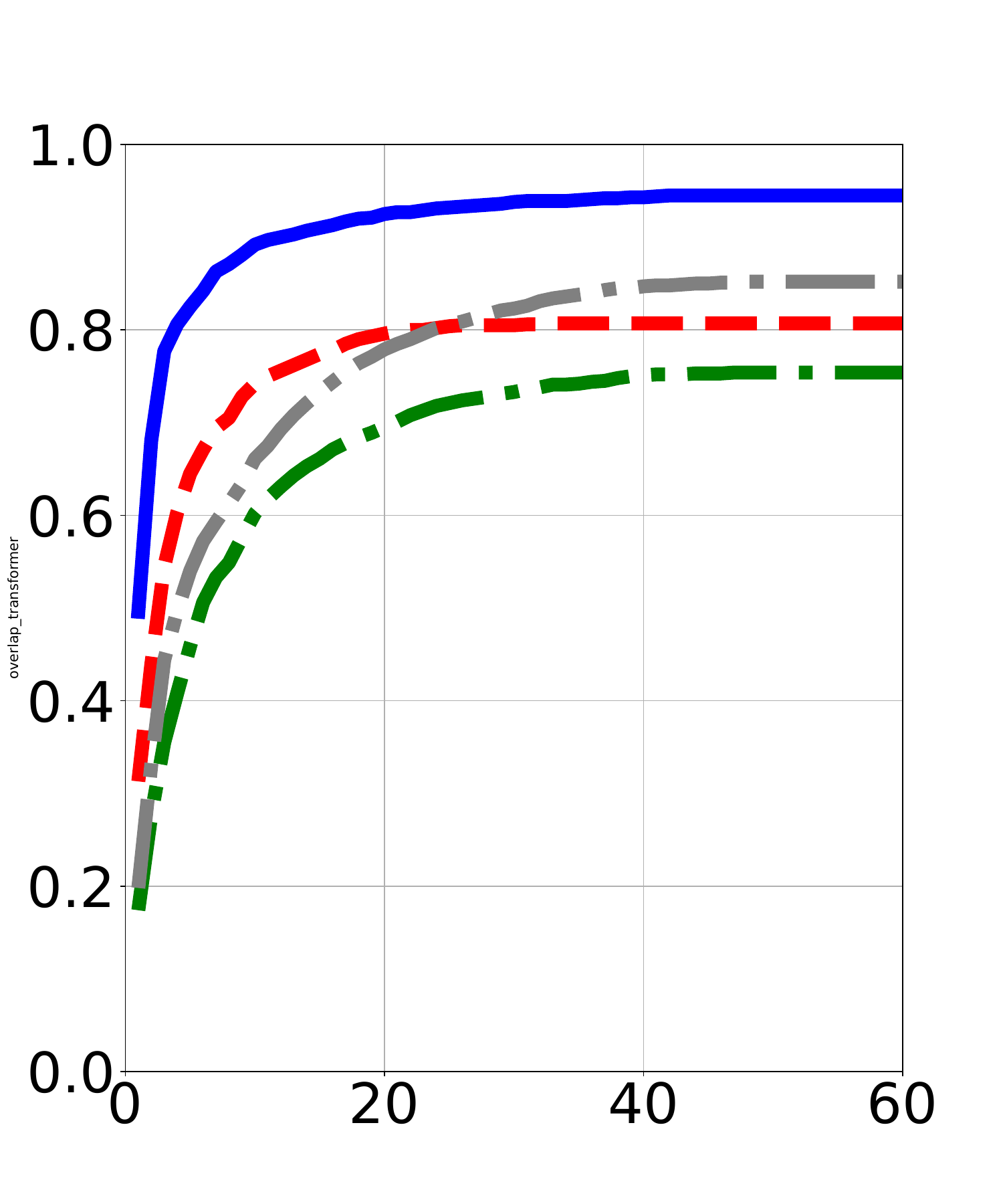}
  \end{subfigure}
  \hfill
  \begin{subfigure}[b]{\figsize\textwidth}
    \centering
    \includegraphics[width=\textwidth, trim={\leftvertical cm \bottom cm 1cm \top cm},clip]{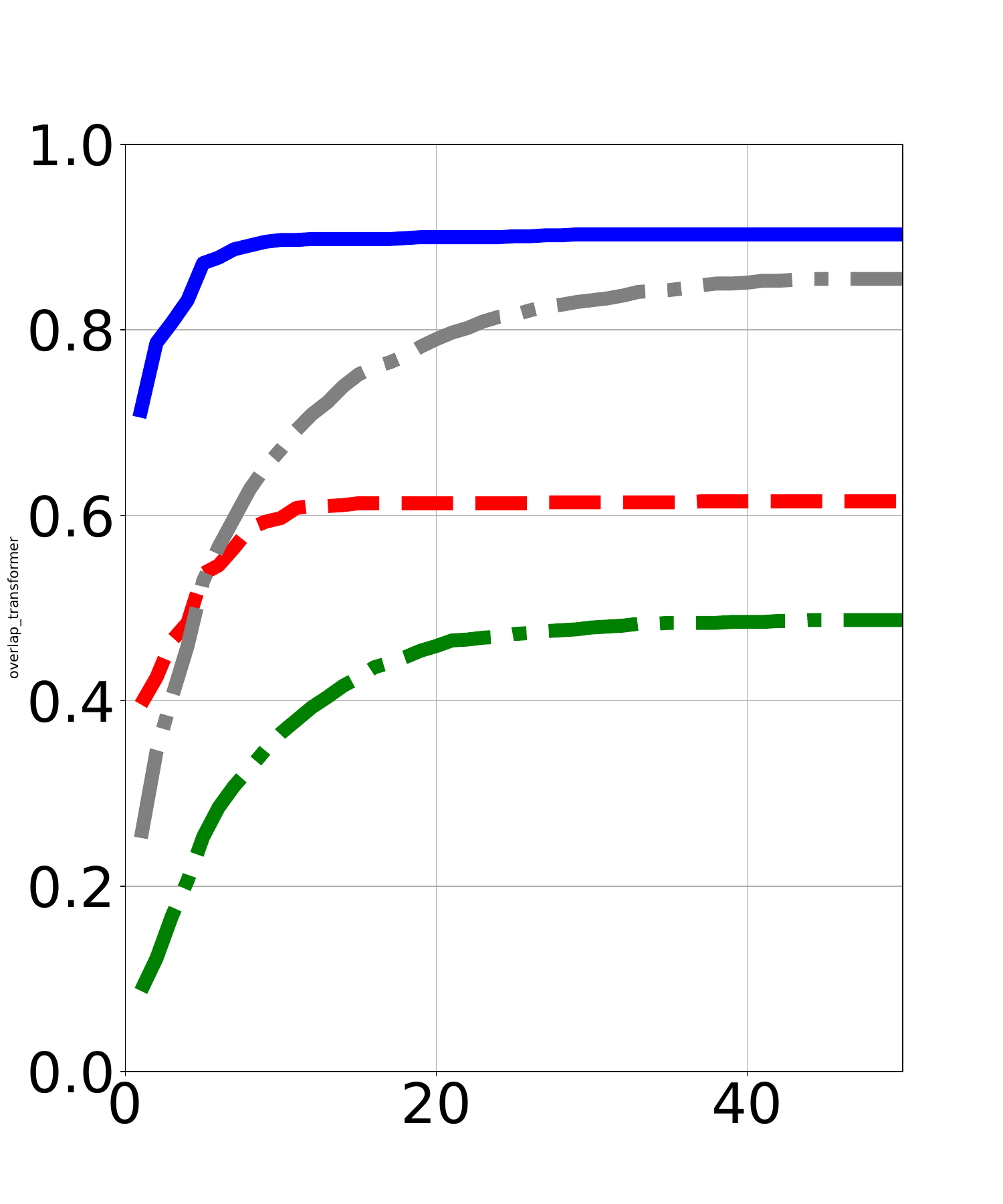}
  \end{subfigure}
  \begin{subfigure}[b]{\figsize \textwidth}
    \centering
    \includegraphics[width=\textwidth, trim={\leftvertical cm \bottom cm 1cm \top cm},clip]{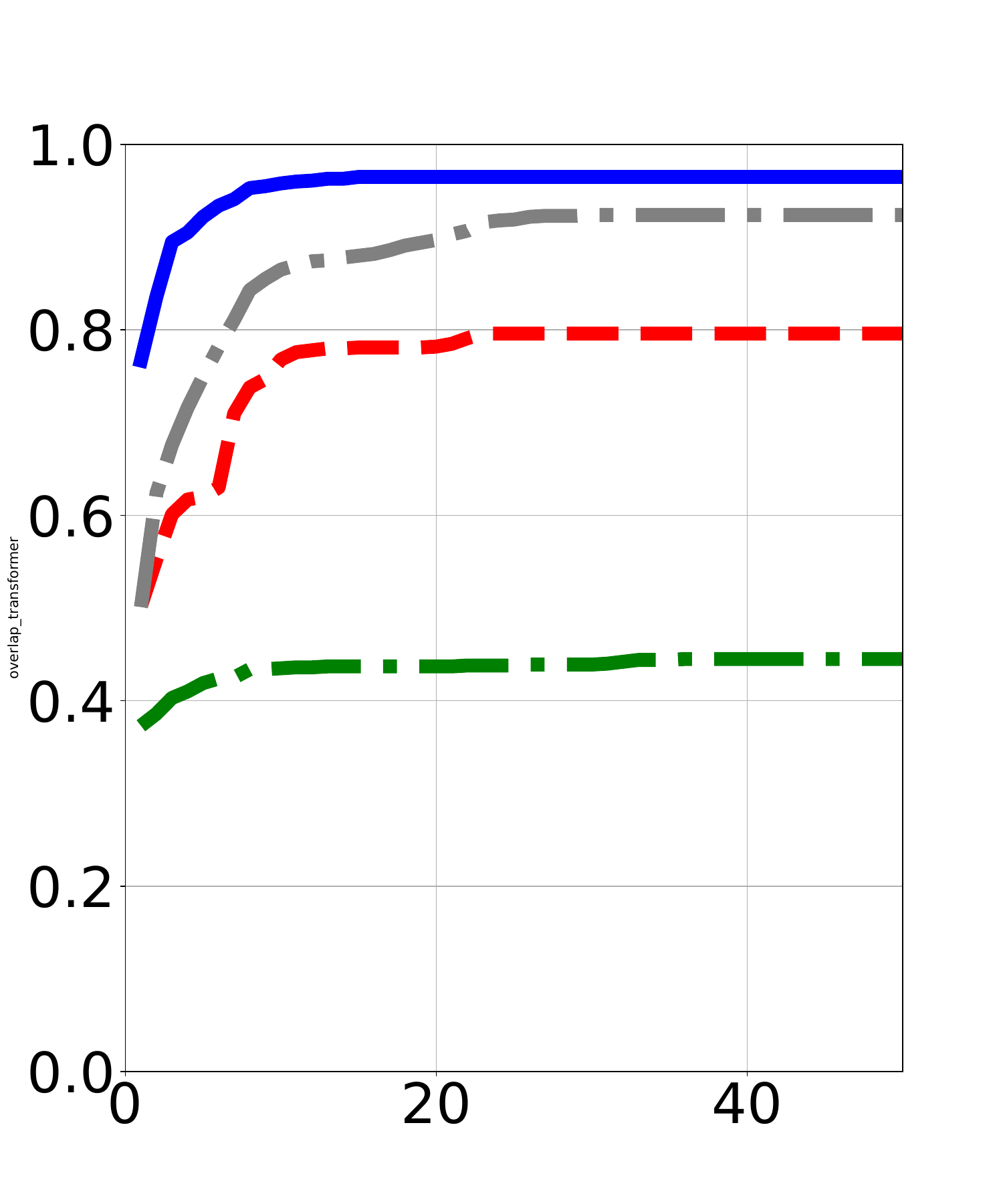}
  \end{subfigure}
  \hfill
  \begin{subfigure}[b]{\figsize\textwidth}
    \centering
    \includegraphics[width=\textwidth, trim={\leftvertical cm \bottom cm 1cm \top cm},clip]{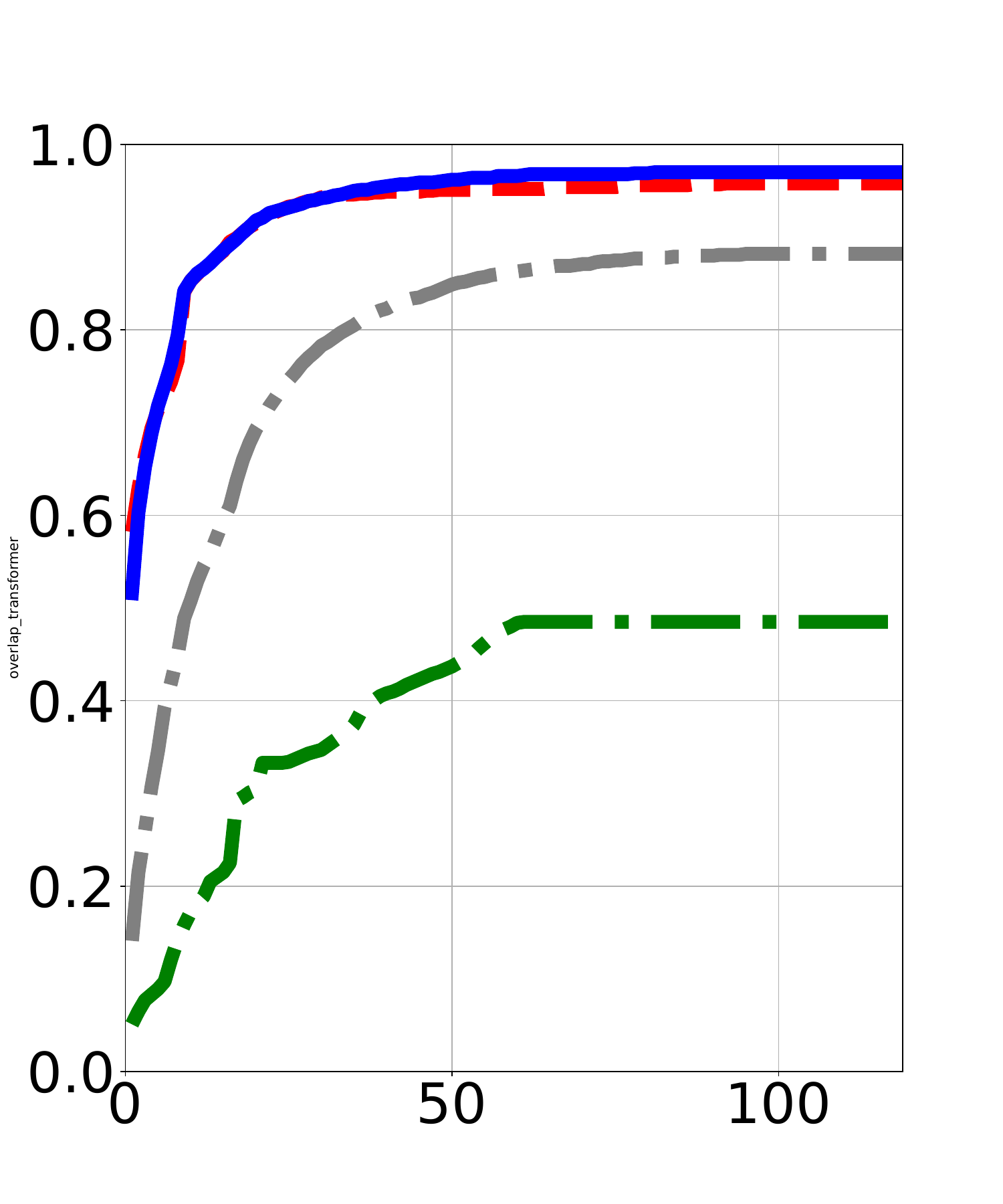}
  \end{subfigure}
   \hfill
  \begin{subfigure}[b]{\textwidth}
    \centering
        \textsize   Range ($r_{th}$) [m]
    \end{subfigure}%
 \hfill
  \begin{subfigure}[b]{0.8\textwidth}
    \centering
    \includegraphics[width=\textwidth]{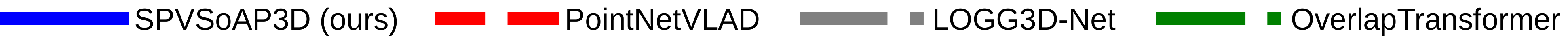}
  \end{subfigure}
  \caption{Performance results along the segments, reported using the Recall@$k$ with $k \in [1,10]$ for $r_{th} \in [1,..,l]$ where $l$ is the segment length.}
  \label{fig:results}
\end{figure*}

\begin{table*}[ht]
\caption{Performance results for $r_{th} = 10\,m$ are reported using Recall@$k$ with $k \in [1, 1\%]$. The bottom four rows refer to results from the ablation study, where the row without any marks represents SPVSoAP3D only with second-order average pooling, \ie, until stage 2 (see Fig. \ref{fig:pipeline}). `LOG' refers to stage 3, `PN' refers to stage 4, while `FC' refers to stage 5.}
{\renewcommand{\arraystretch}{1.2}
\begin{adjustbox}{max width=\textwidth}
\begin{tabular}{llll|ccccccc|ccccccc}
\noalign{\hrule height 1pt}\hline 
\multicolumn{4}{l}{}  & \multicolumn{7}{c|}{Recall@1} & \multicolumn{7}{c}{Recall@1\%} \\ \hline

\multicolumn{4}{l}{}    &ON23&GTJ23& ON22& OJ23&OJ22 &SJ23 & MEAN   &ON23&GTJ23& ON22& OJ23&OJ22 &SJ23 &  MEAN\\ 
\hline

\multicolumn{4}{l|}{PointNetVLAD\cite{angelina2018pointnetvlad}}&0.184&0.649&0.610&0.521&0.686&\textbf{0.659}&0.551&0.652&0.920&0.926&0.754&0.873&0.965&0.848\\
\multicolumn{4}{l|}{OverlapTransformer\cite{9785497}}&0.023&0.073&0.293&0.150&0.302&0.075&0.153&0.300&0.679&0.854&0.487&0.476&0.203&0.500\\
\multicolumn{4}{l|}{LOGG3D-Net\cite{9811753}} &0.074&0.248&0.303&0.182&0.364&0.153&0.221& 0.693&0.947&0.975&0.950&0.989&0.928&0.914\\ \hline \noalign{\hrule height 0.1pt}\hline  \hline \noalign{\hrule height 0.1pt}\hline 

  \multirow{5}{*}{\rotatebox[origin=c]{90}{\scriptsize SPVSoAP3D}} & FC  & LOG  & PN &
  ON23&GTJ23& ON22& OJ23&OJ22 &SJ23 &  MEAN & 
  ON23 & GTJ23 & ON22& OJ23&OJ22 &SJ23 &  MEAN\\  \cline{2-18} 
 &  & & & 
 0.200&0.692&0.467&0.649&\textbf{0.862}&0.501&0.562 &  
 0.826&\textbf{0.997}&0.896&0.978&0.960&0.983&0.940\\  
 & \checkmark & & &
 0.209&0.740&0.676&0.663&0.817&0.484&0.598 &
 0.829&0.981&0.981&0.978&0.935&0.977&0.947\\ 
  & \checkmark    &  \checkmark  &  & 
  0.189&0.721&\textbf{0.707}&0.677&0.820&0.541&0.609&
  0.844&0.966&\textbf{0.989}&\textbf{0.982}&0.952&0.982&0.953\\
  & \checkmark    &  \checkmark   & \checkmark & 
  \textbf{0.268}&\textbf{0.763}&0.674&\textbf{0.704}&0.820&0.554&\textbf{0.630}& 
  \textbf{0.887}&0.966&0.988&0.963&\textbf{0.994}&\textbf{0.986}&\textbf{0.964}\\ 
\noalign{\hrule height 1pt}\hline
\end{tabular}
\end{adjustbox}
}
\label{tab:results}
\end{table*}

\subsection{Results \& Discussion } \label{sec:results}
In this section, we present and discuss the empirical results. As outlined in Section \ref{sec:dep}, we conducted two experiments comparing SPVSoAP to SOTA models, namely PointNetVLAD, OverlapTransformer, and LOGG3D-Net. 

The results of the experiment with a fixed radius ($r_{th}=10\,m$) are reported in Table \ref{tab:results}. This table also includes the results of an ablation study, which highlights the contribution of each stage in the aggregation module to the overall performance. The results of the experiment with $r_{th}$ varying along the segments are reported in Fig. \ref{fig:results}. 

\subsubsection{Comparison to SOTA}
In the experiment with $r_{th} = 10\,m$, the results indicate that SPVSoAP3D, on average, outperforms the SOTA models. It achieves 7.8\,percentage points (pp) higher performance than the next best model in top-1 retrieval (PointNetVLAD), and 5\,pp in top-1\% (LOGG3D-Net). Particularly, when compared to LOGG3D-Net, which shares the same backbone and uses second-order max pooling, the proposed approach performs 40\,pp higher in top-1 and 5\,pp  in top-1\%, suggesting that the average operator is a more suitable aggregation approach for horticultural environments than the max operator. 

Compared to PointNetVLAD and OverlapTransformer, which rely on first-order statistics, SPVSoAP3D exhibits greater robustness, achieving higher performance on average in both top-1 and top-1\% retrieval. SPVSoAP3D only ranks second in the SJ23 sequence  to PointNetVLAD in top-1 retrieval. These results hold true when evaluating along the segments, with all models retrieving more true loops as the range increases along the segments. Upon analyzing the results of recall@10 (second row) in Fig. \ref{fig:results}, SPVSoAP3D surpasses  PointNetVLAD in SJ23, showcasing the superior  modeling capacity of the proposed approach.

\subsubsection{Ablation Study}
This study highlights the contribution of each stage in the aggregation process to the overall performance. The results of this study are presented in the lower section of Table \ref{tab:results}, where FC refers to stage 5 (\ie, Output Layer) in Fig \ref{fig:pipeline}. Regardless of whether FC is utilized, descriptor normalization is always computed. LOG refers to stage 3 (Log-Euclidean projection), while PN refers to stage 4 (power normalization).  The first row of the ablation study section in Table \ref{tab:results} (without any marks) refers to SPVSoAP3D using only second-order average pooling  (up to stage 2, included).

The results suggest that the average operator is sufficient to outperform SOTA models. However, when  all stages are included, retrieval performance increases by 6.8\,pp for top-1 and 2.4\,pp for top-1\%. These findings highlight the effectiveness of the descriptor enhancement strategy proposed in this work.

\section{CONCLUSIONS}
We have introduced SPVSoAP3D, a novel 3D LiDAR-based place recognition model suitable for agricultural robotics, and extend the HORTO-3DLM dataset with two new sequences from horticultural environments: one recorded in an orchard in Metz, France, and another in a greenhouse in Coimbra, Portugal.  This expansion enriches the dataset with more diversified data, enhancing its utility for  agricultural robotics research.

Our proposed SPVSoAP3D model utilizes a second-order average pooling approach and an additional enhancement stage for local feature aggregation. Through experimentation, and compared to SOTA approaches, the results show the superiority of the average operator over the max operator on second-order statistics in horticultural environments. Moreover, SPVSoAP3D outperforms models relying on first-order approaches.  Additionally, the results highlight the effectiveness of the enhancement stage, yielding performance improvements compared to SPVSoAP3D using second-order average pooling alone.

\section*{ACKNOWLEDGMENTS}
This work has been supported by the project GreenBotics (ref. PTDC/EEI-ROB/2459/2021), funded by Fundação para a Ciência e a Tecnologia (FCT), Portugal. It was also partially supported by FCT through grant UIDB/00048/2020 and under the PhD grant with reference 2021.06492.BD. 
Furthermore, we extend our sincere thanks to Cueillette de Peltre for providing access to their orchards.

\bibliographystyle{IEEEtran}
\bibliography{ref}

\end{document}